%% file: bare_jrnl.tex
\documentclass[journal,onecolumn,draftclsnofoot]{IEEEtran}

\usepackage[utf8]{inputenc} 

\ifCLASSINFOpdf
\else
\fi
\usepackage{array,booktabs} 
\usepackage{amsmath,amssymb} 
\usepackage{stfloats} 
\usepackage{hyperref}
\usepackage{color}
\usepackage{flushend}
\usepackage{float}

\usepackage{algorithm,algorithmicx,algpseudocode}

\usepackage{amsfonts,bm} 
\usepackage{siunitx}

\usepackage{graphics} 
\usepackage{graphicx}
\usepackage{epsfig} 
\usepackage{times} 
\usepackage{xcolor}
\usepackage[fleqn]{mathtools}

\input{new_commands.tex}

\begin{document}
%
\title{Lie Algebraic Unscented Kalman Filter for Pose Estimation}
%
%
%

\author{Alexander~M.~Sjøberg,
        Olav~Egeland
\thanks{The authors are with the Department
of Mechanical and Industrial Engineering, Norwegian University of Science and Technology (NTNU), Trondheim, Norway. e-mail: alexander.m.sjoberg@ntnu.no, olav.egeland@ntnu.no.}
}

\maketitle

\begin{abstract}
An unscented Kalman filter for matrix Lie groups is proposed where the time propagation of the state is formulated on the Lie algebra. This is done with the kinematic differential equation of the logarithm, where the inverse of the right Jacobian is used. The sigma points can then be expressed as logarithms in vector form, and  time propagation of the sigma points and the computation of the mean and the covariance can be done on the Lie algebra. The resulting formulation is to a large extent based on logarithms in vector form, and is therefore closer to the UKF for systems in $\mathbb{R}^n$. This gives an elegant and well-structured formulation which provides additional insight into the problem, and which is computationally efficient. The proposed method is in particular formulated and investigated on the matrix Lie group $SE(3)$. A discussion on right and left Jacobians is included, and a novel closed form solution for the inverse of the right Jacobian on $SE(3)$ is derived, which gives a compact representation involving fewer matrix operations. The proposed method is validated in simulations.
\end{abstract}


%
\IEEEpeerreviewmaketitle


\input{Introduction.tex}
\input{Preliminaries.tex}

\input{LieAlgUKF.tex}

\input{ExperimentandSimulation.tex}

\input{Conclusion.tex}

\input{appendix.tex}

\section*{Acknowledgment}
The research presented in this paper was funded by the Norwegian Research Council under Project Number 237896, SFI Offshore Mechatronics.

\ifCLASSOPTIONcaptionsoff
  \newpage
\fi



%

\bibliographystyle{abbrv}
\bibliography{refs}

\end{document}

%% file: new_commands.tex
\newcommand{\bma}{{\bm{a}}}
\newcommand{\bmb}{{\bm{b}}}
\newcommand{\bmc}{{\bm{c}}}
\newcommand{\bmd}{{\bm{d}}}
\newcommand{\bme}{{\bm{e}}}

\newcommand{\bmm}{{\bm{m}}}

\newcommand{\bmr}{{\bm{r}}}
\newcommand{\bms}{{\bm{s}}}

\newcommand{\bmu}{{\bm{u}}}
\newcommand{\bmv}{{\bm{v}}}
\newcommand{\bmw}{{\bm{w}}}

\newcommand{\bmy}{{\bm{y}}}

\newcommand{\bmC}{{{\bm C}}}

\newcommand{\bmI}{{{\bm I}}}
\newcommand{\bmJ}{{{\bm J}}}
\newcommand{\bmK}{{{\bm K}}}

\newcommand{\bmN}{{{\bm N}}}

\newcommand{\bmP}{{{\bm P}}}
\newcommand{\bmQ}{{{\bm Q}}}
\newcommand{\bmR}{{{\bm R}}}

\newcommand{\bmT}{{{\bm T}}}

\newcommand{\bmV}{{{\bm V}}}

\newcommand{\bmX}{{{\bm X}}}
\newcommand{\bmY}{{{\bm Y}}}
\newcommand{\bmZ}{{{\bm Z}}}
\newcommand{\bmxi}{{\bm \xi}}
\newcommand{\bmeta}{{\bm \eta}}

\newcommand{\bmsigma}{{\bm \sigma}}

\newcommand{\bmtheta}{{\bm \theta}}

\newcommand{\bmPhi}{{\bm \Phi}}
\newcommand{\bmPsi}{{\bm \Psi}}
\newcommand{\bmrho}{{\bm \rho}}

\newcommand{\bmzeta}{{\bm \zeta}}
\newcommand{\bmomega}{{\bm \omega}}

\newcommand{\bmnu}{{\bm \nu}}
\newcommand{\bmmu}{{\bm \mu}}
\newcommand{\bmepsilon}{{\bm \epsilon}}

\newcommand{\bal}{\begin{aligned}}
\newcommand{\eal}{\end{aligned}}
\newcommand{\bmat}{\begin{bmatrix}}
\newcommand{\emat}{\end{bmatrix}}
\newcommand{\bcmat}{\begin{Bmatrix}}
\newcommand{\ecmat}{\end{Bmatrix}}
\newcommand{\bpmat}{\begin{pmatrix}}
\newcommand{\epmat}{\end{pmatrix}}

\newcommand{\tr}{{\textrm{T}}}

\newcommand{\inv}{{-1}}

\newcommand{\ad}{{\textbf{ad}}}

\newcommand{\Chol}{{\mathrm{Chol}}}

%% file: Introduction.tex

\section{Introduction}

The use of Lie group theory for attitude and pose estimation has received considerable attention in the research literature. The reason for this is that the set of rotation matrices $SO(3)$ and the set of homogeneous transformation matrices $SE(3)$ are both matrix Lie groups. Matrix Lie groups have a number of properties that are useful in the design of estimators and observers. In addition, unit quaternions form a Lie group, and some of the design techniques for quaternion estimators and observers can be related to their Lie group properties. The main branches of methods for Lie group estimators are based on Kalman filtering and nonlinear observer design. 

Early work on nonlinear attitude estimation and control with quaternions is found in \cite{Salcudean1991}, \cite{Wen1991} and \cite{Egeland1994}, where the structural properties of the unit quaternions were used in the design. This was used in navigation for pose estimation in \cite{Vik2001} and for attitude estimation in \cite{Thienel2003}, where bias estimation was included. An important development in Kalman filtering based on quaternions was the multiplicative extended Kalman filter \cite{lefferts1982} where the global attitude was represented by the 4-dimensional unit quaternion, while the 3-dimensional quaternion vector was estimated at each time step. This was later generalized to alternative 3-dimensional vector representations of attitude, including the modified Rodrigues parameters \cite{Crassidis2007} and the rotation vector \cite{Pittelkau2003}, which is the vector form of the logarithm in $SO(3)$ \cite{Park1995}. An unscented Kalman filter (UKF) \cite{JulierUhlmann1997} was developed for attitude estimation in \cite{Crassidis2003}, where the kinematic differential equation for the quaternions was used for the time propagation of the sigma points. Another example of attitude estimation on quaternions using the UKF is found in \cite{Sipos2008}. The multiplicative extended Kalman filter for quaternions have been extended to pose estimation by introducing dual quaternions \cite{Tsiotras2015}, while in \cite{Deng2016} a UKF was developed for pose estimation using dual modified Rodrigues parameters.  

An important development was the nonlinear complementary filter \cite{Mahony2008}, which was an attitude observer where the global attitude was represented by a rotation matrix, and the observation error was represented by the 3-dimensional vector form of the anti-symmetric part of the rotation matrix. The resulting filter is robust and well suited for low-cost sensors, and for a number of different sensor configurations. The nonlinear complementary filter was generalized to the special linear group in \cite{Mahony2012}, and to $SE(3)$ in \cite{Baldwin2007}. A related work on pose estimation is found in \cite{Rehbinder2003}, where vision and inertial sensors are used. A nonlinear observer for Lie groups based on Riemannian gradient descent was presented in \cite{Lageman2010,Hua2011}. This work also addressed the invariance properties that were addressed in the symmetry-preserving observer of \cite{Bonnabel2008,Bonnabel2009}, which was further developed to an invariant extended Kalman filter that was used as a stable observer on Lie groups \cite{Barrau2017}, and for consistency in extended Kalman filtering for SLAM \cite{BrossardExploiting2019}.   

The concept of a concentrated Gaussian distribution on Lie groups was introduced in \cite{Wang2006,Chirikjian2012}, where a normal distribution on a Lie group was defined in terms of a normal distribution of the logarithm in vector form.  This was further developed in \cite{Barfoot2014}, where this formulation was used for fusion of multiple measurements of pose. The formulation of \cite{Barfoot2014} was used in \cite{Bourmaud14} to formulate an extended Kalman filter (EKF) for matrix Lie groups where the covariance was calculated for the concentrated Gaussian distribution, and the time propagation of the state and the covariance was derived from the first order approximation of the Baker-Campbell-Hausdorff (BCH) formula. A similar approach was used in \cite{Cesic2017a} for an extended information filter on matrix Lie groups. The method of \cite{Bourmaud14,Bourmaud13} was further developed in \cite{Sjoberg2019} where the time propagation of the state and the covariance of an EKF was computed on the Lie algebra using the kinematic differential equations of the logarithm.   

In \cite{Hauberg2013} a UKF was formulated for Riemannian manifolds. This was done by generating sigma points as elements of the manifold, and then calculating the mean by minimization on the manifold, while the covariance was calculated in the tangent plane of the mean. It was remarked that the sigma points and the mean can alternatively be calculated in the tangent plane. In \cite{Hertzberg2013} a UKF framework for sensor fusion on manifolds was presented where the sigma points were given on the manifold. A UKF for quadrotors on $SE(3)$ was presented in \cite{Loianno2016}, where the sigma points were computed on the manifold. In \cite{Brossard2017} the concept of concentrated Gaussian distributions was used to formulate a UKF for Lie groups where the sigma points are in the Lie algebra, while the time propagation is formulated on the Lie group. In \cite{Menegaz2018} a systematic overview of Riemannian extensions of UKFs was presented based on the formalism in \cite{Menegaz2015}. A simulation study was included where a UKF was implemented for unit quaternions, where the sigma points were computed on the manifold, and the mean was found as an optimization problem on the manifold. The paper stated that future work should focus on computationally efficient UKFs for Riemannian manifolds for real-time applications. In \cite{Magalheas2018} a UKF is formulated where the sigma points are calculated in the Lie algebra, while the time propagation was in the manifold.   

The main contribution of the present paper is that the time propagation is formulated in terms of the kinematic differential equation of the logarithm, using the inverse of the right Jacobian. Then time propagation of the sigma points can be formulated on the Lie algebra, and moreover, the mean and the covariance can be computed on the Lie algebra. The covariance matrix is transformed between different tangent planes based on the BCH formula using the right Jacobian. The measurement update is based on \cite{Brossard2017}. The resulting formulation is to a large extent given in terms of logarithms in vector form, which makes the proposed UKF more similar to the original formulation for $\mathbb{R}^n$. This may lead to added insight and ease of implementation. Moreover, some of the steps of the method will be computationally efficient. In particular, time propagation of the group element and optimization on the manifold is avoided, and the method involves few calculations of exponentials and logarithms. The right Jacobian is important in our method as it appears in the kinematic differential equation of the logarithm. A new closed form solution for the inverse right and left Jacobian in $SE(3)$ is derived in the paper based on \cite{Bullo1995, Barfoot2014}.

The paper is organized as follows. Section 2 presents basic theory on Lie groups and Lie algebras including probability distributions and the left and right Jacobians. In Section 3 a new closed form solution for the right and left Jacobian in $SE(3)$ is derived. Then in Section 4 a Lie Algebraic UKF on $SE(3)$ is presented. Finally, the performance of the proposed UKF is demonstrated in simulations.


%% file: Preliminaries.tex
\section{Preliminaries} \label{sec:preliminaries}

\subsection{Matrix Lie groups}
Let $G$ be a matrix Lie group, and let $\mathfrak{g}$ be the associated Lie algebra \cite{Hall2003,Chirikjian2012}. Consider the exponential of $\mathfrak{u} \in \mathfrak{g}$, which is 
\begin{align}\label{eq:exponetialTaylorSeriesExpansion}
    \bmX = \exp \mathfrak{u} = \sum^\infty_{k=0}\frac{\mathfrak{u}^k}{k!} \in G
\end{align}
It follows that $\mathfrak{u}$ is the logarithm of $\bmX$, which is written
\begin{equation}
\mathfrak{u} = \log \bmX 
\end{equation}
An element $\mathfrak{u}\in \mathfrak{g}$ of the Lie algebra can be can be represented by the vector $\bmu = [u_1,\ldots, u_n]^\tr \in \mathbb{R}^n$. The notation $[\mathfrak{u}]^\vee_G = \bmu\in \mathbb{R}^n$ and $[\bmu]^\wedge_G = \mathfrak{u}\in \mathfrak{g}$ is used in agreement with \cite{Bourmaud14}. 

Let $\mathfrak{a},\mathfrak{b} \in \mathfrak{g}$ be elements of the Lie algebra with vector representations $\bma = [\mathfrak{a}]_G^\vee$ and $\bmb = [\mathfrak{b}]^\vee_G$. Then the adjoint map $ad_G(\mathfrak{a})$ and its matrix form $\textbf{ad}_G(\bma)\in \mathbb{R}^{n\times n}$ are given by 
\begin{equation}
[\textbf{ad}_G(\bma)\bmb]^\wedge_G = ad_G(\mathfrak{a})\mathfrak{b} = [\mathfrak{a}, \mathfrak{b}]
\end{equation}
where $[\mathfrak{a}, \mathfrak{b}] = \mathfrak{a}\mathfrak{b} - \mathfrak{b} \mathfrak{a}$ is the Lie bracket.   

The kinematic differential equation for $\bmX\in G$ is given by   
\begin{align}\label{KinematicDiffEqX}
    \dot\bmX = [\bmv_l]_G^\wedge \bmX = \bmX [\bmv_r]_G^\wedge
\end{align}
where $\bmv_l\in \mathbb{R}^n$ is the vector form of the left velocity and $\bmv_r\in \mathbb{R}^n$ is the vector form of the right velocity. 

There is an alternative form of the kinematic differential equation which is formulated in terms of the  logarithm. This is found from the time derivative of the exponential function $\bmX(t) = \exp([\bmu(t)]_G^\wedge)$, which is \cite{Faraut2008}
\begin{equation}\label{kinematicDiffEqXLogarithmR}
    \dot\bmX = [\bmJ_l(\ad(\bmu)) \dot\bmu]_G^\wedge\bmX = \bmX [\bmJ_r(\ad(\bmu)) \dot\bmu]_G^\wedge
\end{equation}
Here $\bmJ_l$ is the left Jacobian and $\bmJ_r$ is the right Jacobian, which are given by  
\begin{align}\label{eq:right_jacobian_definitionR}
    \bmJ_l(\ad(\bmu)) &= \bmJ_r(-\ad(\bmu)) = \sum_{i = 0}^\infty\frac{(\ad(\bmu))^i}{(i+1)!}  
\end{align}
From (\ref{KinematicDiffEqX}) and (\ref{kinematicDiffEqXLogarithmR}) is it seen that the kinematic differential equation for the logarithm can be written in vector form as  \cite{Bullo1995} 
    \begin{equation}\label{kinematicDifferentialEqLogarithmR}
    \dot\bmu = \bmJ_l^\inv(\ad(\bmu)) \bmv_l = \bmJ_r^\inv(\ad(\bmu)) \bmv_r 
    \end{equation}
The inverse of the left and right Jacobian is
\begin{equation}\label{eq:inv_right_jacobian_definition}
    \bmJ_l^\inv(\ad(\bmu)) = \bmJ_r^\inv(-\ad(\bmu)) = \sum_{i=0}^\infty\frac{B_i (\ad(\bmu))^i}{i!} 
\end{equation} 
where $B_n$ are the Bernoulli numbers $B_0 = 1$, $B_1 = -\frac{1}{2}$, $B_2 = \frac{1}{6}$, $B_3 = 0$, $B_4 = -\frac{1}{30}$, $B_5 = 0$,\ldots.

\subsection{The Baker-Campbell-Hausdorff formula}

Consider the elements $\mathfrak{a} = [\bma]^\wedge_G,\mathfrak{b} = [\bmb]^\wedge_G,\mathfrak{c} = [\bmc]^\wedge_G$ of the Lie algebra $\mathfrak{g}$, and suppose that 
\begin{equation}\label{eq:exp(c)=exp(a)exp(b)}
    \exp(\mathfrak{c}) = \exp(\mathfrak{a})\exp(\mathfrak{b}).
\end{equation} 
Then, according to the Baker-Campbell Hausdorff (BCH) formula \cite{Hall2003},
\begin{equation}
    \mathfrak{c} =  \mathfrak{a} +  \mathfrak{b} + \frac{1}{2} [\mathfrak{a},\mathfrak{b}] 
    + \frac{1}{12} [\mathfrak{a},[\mathfrak{a},\mathfrak{b}]] + \frac{1}{12} [\mathfrak{b},[\mathfrak{b},\mathfrak{a}]] + \ldots
\end{equation}
If only first order terms of $\mathfrak{b}$ are included, then the vector representation $\bmc$ can be approximated as \cite{Klarsfeld1989}
\begin{equation}\label{BCH-c=a+Jrinvb}
    \bmc =  \bma+ \bmJ_r^\inv(\textbf{ad}(\bma))\bmb 
\end{equation}
It is noted that if $\bmc = \bma + \bmd$, then it follows from \eqref{BCH-c=a+Jrinvb} that $\bmd = \bmJ_r^\inv(\textbf{ad}(\bma))\bmb$. This leads to the two first order approximations
\begin{align}
    \exp([\bma]^\wedge_G)\exp([\bmb]^\wedge_G) 
    &=  \exp([\bma + \bmJ_r^\inv(\textbf{ad}(\bma))\bmb]^\wedge_G) 
    \label{BCHExpEquation1}\\
     \exp([\bma + \bmd]^\wedge_G) 
     &= \exp([\bma]^\wedge_G)\exp([\bmJ_r(\textbf{ad}(\bma))\bmd]^\wedge_G) 
     \label{BCHExpEquation2}
\end{align}
which were used in \cite{Barfoot2014}, \cite{Bourmaud13} and \cite{Cesic2017a}.

\subsection{Random Variables and Concentrated Gaussian Distributions on Matrix Lie Groups}

A random variable $\bmX \in G$ is said to have the normal distribution  $\mathcal{N}_G(\bar\bmX, \bmP)$ on $G$ if \cite{Barfoot2017}
\begin{align}\label{eq:cgd_right}
    \bmX = \bar\bmX \exp([\bmu]^\wedge_G)
\end{align}
where the vector form of the logarithm 
\begin{align}
    \bmu \sim \mathcal{N}_{\mathbb{R}^n}(\bm0, \bmP)
\end{align}
is normally distributed with zero mean. It is required that the distribution is tightly focused around $\bar\bmX$. 

Next, consider a random variable $\bmY = \exp([\bmxi]^\wedge_G) \in G$, where $\bmxi \sim \mathcal{N}_{\mathbb{R}^n}(\bar{\bmxi}, \bmQ)$ is normally distributed with nonzero mean. The zero-mean vector $\delta\bmxi = \bmxi - \bar{\bmxi} \sim \mathcal{N}_{\mathbb{R}^n}(\bm0, \bmQ)$ is introduced. Then from (\ref{BCHExpEquation2}) it is seen that a first order approximation in $\delta\bmxi$ is given by 
\begin{align}\label{exp_xi_with_nonzero_mean_BCH}
    \bmY
    &= \exp([\bar{\bmxi}+\delta\bmxi]^\wedge_G)
    = \exp([\bar{\bmxi}]^\wedge_G) \exp([\bme]^\wedge_G)
\end{align}
where 
\begin{equation}
    \bme = \bmJ_r(\ad(\bar\bmxi))\delta\bmxi \sim \mathcal{N}_{\mathbb{R}^n}(\bm0, \bmR)
\end{equation}
is normally distributed with zero mean and covariance
\begin{align}
\bmR &= \bmJ_r(\ad(\bar{\bmxi})) \bmQ \bmJ_r^\tr(\ad(\bar{\bmxi}))
\end{align}
It follows that 
\begin{align}
 \bmY = \bar\bmY\exp([\bme]^\wedge_G)  \sim \mathcal{N}_G(\bar\bmY, \bmR)
\end{align}
where $\bar\bmY = \exp([\bar{\bmxi}]^\wedge_G)$. 
This result was derived in \cite{Bourmaud14} for use in the update of an extended Kalman filter on a matrix Lie group. A related problem was treated in \cite{Cesic2017b} where a merging algorithm for Gaussian components on $G$ was developed.

\subsection{Calculation of mean and covariance}

Consider a set 
\begin{equation}
  \bmY_i = \exp([\bmxi_i]^\wedge_G) \in G  
\end{equation}
of $N$ Lie group elements with corresponding logarithms given in vector form by $\bmxi_i$. In \cite{Hauberg2013} the elements were expressed in terms of the mean $\bmmu\in G$ as 
\begin{equation}
    \bmY_i = \bmmu\exp([\bmepsilon_i]_G^\wedge)
\end{equation}
where the mean was found from the minimization problem 
\begin{equation}\label{Hauberg_mean_minimization}
    \bmmu = \arg\min_{\bmmu\in G} d(\bmY_i,\bmmu)^2
\end{equation}
for some distance function $d$. The empirical covariance was calculated from  
\begin{equation}
    \bmP_G = \frac{1}{N}\sum_{i=1}^N \bmepsilon_i\bmepsilon_i^\tr
\end{equation}
where $[\bmepsilon_i]^\wedge_G = \log(\bmmu^\inv\bmY_i)$.

The calculation of the mean as the minimization problem (\ref{Hauberg_mean_minimization}) on $G$ may be time consuming in real time applications. Therefore, we suggest that the mean is calculated on the tangent space as
\begin{equation}\label{calculate_mean_from_average_logarithm}
    \bar\bmY = \exp([\bar{\bmxi}]^\wedge_G)
\end{equation}
where 
\begin{equation}\label{calculate_average_logarithm}
    \bar{\bmxi} = \frac{1}{N} \sum_{i=1}^N \bmxi_i
\end{equation}
It is seen from (\ref{exp_xi_with_nonzero_mean_BCH}) that this leads to the first order approximation 
\begin{equation}
    \bmY_i = \bar\bmY\exp(\bme_i)
\end{equation}
where $\bme_i = \bmJ_r(\ad(\bar{\bmxi}))\delta\bmxi_i$ where $\delta\bmxi_i = (\bmxi_i-\bar{\bmxi})$. The empirical covariance can then be calculated as 
\begin{equation}
    \bmP = \frac{1}{N}\sum_{i=1}^N \bme_i\bme_i^\tr
    = \bmJ\left(\frac{1}{N}\sum_{i=1}^N \delta\bmxi_i\delta\bmxi_i^\tr\right)\bmJ^\tr
\end{equation}
where $\bmJ = \bmJ_r(\ad(\bar{\bmxi}))$. 

\subsection{Calculation of mean by optimization}\label{sec:calc_of_mean_manton}

Consider the case where the distance function (\ref{Hauberg_mean_minimization}) is 
\begin{equation}\label{distance_Riemannian_metric}
  d(\bmY_i,\bmmu)^2 = \bmepsilon_i^\tr\bmepsilon_i
\end{equation}
which is the usual angular distance in $SO(3)$ and a left-invariant metric in $SE(3)$. Then the mean calculated by minimization on the group will be
\begin{equation}\label{minimization_mu}
    \bmmu = \arg\min_{\bmmu\in G} [\log(\bmmu^\inv\bmY_i)]^\vee_G)^\tr[\log(\bmmu^\inv\bmY_i)]^\vee_G 
\end{equation}
In comparison to this, the calculation of the mean $\bar\bmY$ by calculating the average logarithm $[\bar{\bmxi}]^\wedge_G$ in (\ref{calculate_average_logarithm}) corresponds to the minimization problem 
\begin{equation}\label{minimization_barY}
    \bar\bmY = \arg\min_{\bar\bmY\in G} ([\log({\bar\bmY}^\inv\bmY_i)]^\vee_G)^\tr\bmJ^\tr\bmJ
    [\log({\bar\bmY}^\inv\bmY_i)]]^\vee_G
\end{equation}
on $G$ where $\bmJ = \bmJ_r(\ad(\bar{\bmxi}))$. It is seen that the only difference between the two optimization problems (\ref{minimization_mu}) and (\ref{minimization_barY}) is the weighting matrix $\bmJ^\tr\bmJ$ in (\ref{minimization_barY}). For small $\bar{\bmxi}$ it is seen from (\ref{eq:right_jacobian_definitionR}) that this weighting matrix will be close to the identity matrix, and it is reasonable to expect that $\bar\bmY$ will be close to $\bmmu$ for this distance function.

It is interesting to note that the calculation of the mean according to (\ref{calculate_mean_from_average_logarithm}) and (\ref{calculate_average_logarithm}) is related to the optimization algorithm of Manton \cite{Manton2004}, who presented a globally convergent numerical algorithm for computing the center of mass on compact Lie groups. This method minimizes the function $f(\bmX) = \frac{1}{2N}\sum_{i=1}^N d(\bmY_i,\bmX)^2$ for $\bmX\in G$ where $\bmY_1,\ldots,\bmY_N \in G$ and $d(\cdot,\cdot)$ is the Riemannian distance function on $G$. The result of the minimization is the Karcher mean. 
The optimization was shown to be globally convergent for compact Lie groups, and was done with a gradient descent method given by
\begin{equation}
    \bmX := \bmX\exp(\mathfrak{a})
\end{equation}
where 
\begin{equation}
    \mathfrak{a} = \frac{1}{N}\sum_{i=1}^N\log(\bmX^\inv\bmY_i) \in g
\end{equation}
It was commented in \cite{Manton2004} that the Lie algebra serves as a first order approximation of the Lie group about the identity, and that the mean on the Lie algebra will approximate the mean on the Lie group with distance function (\ref{distance_Riemannian_metric}). Moreover, it is seen that the calculation of the mean on the Lie algebra with (\ref{calculate_mean_from_average_logarithm}) and (\ref{calculate_average_logarithm}) corresponds to the first step of Manton's method with initial value $\bmX = \bmI$. In the following we will propose a UKF on $G$ where we use  (\ref{calculate_mean_from_average_logarithm}) and (\ref{calculate_average_logarithm}) to calculate the mean and the associated covariance of the sigma points on the Lie algebra.  

\subsection{Time integration and discrete-time model}

The two alternative formulations (\ref{KinematicDiffEqX}) and (\ref{kinematicDifferentialEqLogarithmR}) of the kinematic differential equations can be discretized with Euler's method from time instant $t_k$ to $t_{k+1} = t_k + h$ where $h$ is the time step, and $\bmv_r$ is assumed to be constant over the time step. Then the differential equation (\ref{KinematicDiffEqX}) for the group element $\bmX$ gives
\begin{align}\label{DiscretizationX}
    \bmX(t_{k+1}) = \bmX(t_{k})\exp(h \bmv_{r}(t_k)) 
\end{align}
while the differential equation (\ref{kinematicDifferentialEqLogarithmR}) for the logarithm $\bmu$ gives 
\begin{align}\label{Discretizationu}
    \bmu(t_{k+1}) = \bmu(t_{k}) + h  \bmJ_r^\inv(\ad(\bmu(t_{k})))\bmv_{r}(t_k) 
\end{align}
Suppose that $\bmX(t_{k}) = \exp([\bmu(t_{k})]^\wedge_G)$. Then it follows from (\ref{BCHExpEquation1}) that
\begin{align}
    \bmX(t_{k+1}) = \exp([\bmu(t_{k}) + h  \bmJ_r^\inv(\ad(\bmu(t_{k})))\bmv_{r}(tk)]^\wedge_G )
\end{align}
is a first order approximation of (\ref{DiscretizationX}). This means that the discretization (\ref{DiscretizationX}) and the discretization (\ref{Discretizationu}) give the same result to the first order. We will use the discretization (\ref{Discretizationu}) in the following to formulate the update equations in the unscented Kalman filter on the matrix Lie group $G$.

Additional results on integration schemes based on (\ref{kinematicDifferentialEqLogarithmR}) are found in \cite{Iserles2005} and \cite{Sveier2019}.

\section{Calculation of inverse Jacobian}

In our proposed UKF for matrix Lie groups the inverse of the right Jacobian plays an important role. We will therefore take a closer look at expressions for the inverse of the right Jacobian, and a novel closed form solution for $SE(3)$ will be developed.

\subsection{Jacobians in SO(3)}

The logarithm in $SO(3)$ is given by $\hat{\bmtheta} \in so(3)$, where $\hat{\bma}$ denotes the skew symmetric form of a vector $\bma\in \mathbb{R}^3$. The matrix form of the adjoint map in $SO(3)$ is $\ad(\bmtheta) = \hat{\bmtheta}$. The rotation matrix is given by the exponential as \cite{Park1995}
\begin{equation}
 \bmR = \exp\hat{\bmtheta} = \bmI + \frac{\sin\theta}{\theta}\hat{\bmtheta} 
 + \frac{(1-\cos\theta)}{\theta^2}\hat{\bmtheta}^2
\end{equation}
where $\theta = \|\bmtheta\|$. 
The right Jacobian in $SO(3)$ and its inverse are given in closed form as \cite{Bullo1995}
\begin{align}
    \bmPsi_r(\hat{\bmtheta}) =& \ \bmI - \frac{1 - \cos\theta}{\theta^2}\hat{\bmtheta} + \frac{\theta - \sin\theta}{\theta^3}\hat{\bmtheta}^2\label{eq:rightjac_so3} \\
    \bmPsi_r^{-1}(\hat{\bmtheta}) =& \ \bmI  +   \frac{1}{2}\hat{\bmtheta} + \frac{1-\frac{\theta}{2}\cot\frac{\theta}{2}}{\theta^2}\hat{\bmtheta}^2
    \label{eq:invrightjac_so3}
\end{align}
The coefficients of the exponential, the Jacobian and the inverse Jacobian are well defined for all $\bmtheta$, which is verified by Taylor series expansion of the coefficients. 


\subsection{Jacobians in SE(3)}
The logarithm of 
\begin{equation}
    \bmT = \bmat \bmR & \bmr \\ \bm0^\tr & 1 \emat \in SE(3) 
\end{equation}
is given by
\begin{equation}
    [\bmxi]_{SE(3)}^\wedge = \bmat \hat{\bmtheta} & \bmrho \\ \bm0^\tr & 0 \emat \in se(3)
\end{equation}
where $\hat{\bmtheta} \in so(3)$, $\bmrho \in \mathbb{R}^3$ and $\bmxi = [\bmtheta^\tr,\bm\rho^\tr]^\tr$ is the vector form of the logarithm. The exponential map is given in closed form and can be computed from \cite{Bullo1995,Park1995}
\begin{equation}\label{eq:expmSE3_CF}
   \bmT =  \exp([\bmxi]_{SE(3)}^\wedge) = \bmat \exp\hat\bmtheta & \bmPsi_l(\hat\bmtheta)\bmrho \\ \bm0^\tr & 1\emat
\end{equation}
where $\exp\hat\bmtheta$ is the exponential function in $SO(3)$, and $\bmPsi_l(\hat\bmtheta)$ is the left Jacobian in $SO(3)$. The logarithm can be computed from \cite{Iserles2005}
\begin{align}\label{eq:logmSE3_CF}
    \hat\bmtheta &= \frac{\sin^\inv\|\bmy\|}{\|\bmy\|}\hat\bmy, \quad 
    \hat\bmy = \frac{1}{2}(\bmR-\bmR^\tr)\\ \nonumber
    \bmrho &= \bmPsi_l^{-1}(\hat{\bmtheta})\bmr
\end{align}
The matrix form of the adjoint map in $SE(3)$ is given by
\begin{align}\label{AdjointMapMatrixSE3}
    \textbf{ad}(\bmxi) = \bmat \hat{\bmtheta} & \bm0 \\ \hat{\bmrho} & \hat{\bmtheta} \emat
\end{align}
The kinematic differential equation in terms of $\bmT$ is given by 
\begin{equation}
    \dot{\bmT} = [\bmV_l]_{SE(3)}^\wedge\bmT = \bmT[\bmV_r]_{SE(3)}^\wedge
\end{equation}
where $[\bmV_l]_{SE(3)}^\wedge \in so(3)$ is the left velocity, and $[\bmV_r]_{SE(3)}^\wedge \in so(3)$ is the right velocity, which have vector forms 
\begin{align}\label{eq:angvel_and_linvel_stacked}
    \bmV_l = \bmat \bmomega_l \\ \bmv_l \emat, \quad \bmV_r = \bmat \bmomega_r \\ \bmv_r \emat
\end{align}
where $\bmomega_l = \bmR \bmomega_r$ and $\bmv_l = \bmR\bmv_r + \hat{\bmr}\bmR\bmomega$.  
It follows from \eqref{kinematicDifferentialEqLogarithmR} that the kinematic differential equation in terms of the logarithm is
\begin{align}\label{eq:kinDE_vr}
    \dot{\bmxi} = \bmPhi^{-1}_l(\ad(\bmxi))\bmV_l = \bmPhi^{-1}_r(\ad(\bmxi))\bmV_r
\end{align}
where $\bmPhi_l$ is the left Jacobian and $\bmPhi_r$ is the right Jacobian in $SE(3)$.

The right Jacobian is given in closed form as \cite{Bullo1995}
\begin{align}\label{eq:barfoot_2014}
    \bmPhi_r(\ad(\bmxi)) = \bmat \bmPsi_{r}(\hat{\bmtheta}) & \bm0 \\ \bmQ_r^\tr & \bmPsi_{r}(\hat{\bmtheta}) \emat
\end{align}
where a closed form solution for the submatrix $\bmQ_r$ was found in \cite{Barfoot2014} to be
\begin{align}
    \bmQ_r =& \sum_{k=1}^\infty \frac{(-1)^k}{(k+1)!} \sum_{i=0}^{k-1}\hat{\bmtheta}^{k-i-1}\hat{\bmrho}\hat{\bmtheta}^{i} \nonumber\\ 
    =& \frac{1}{2}\hat\bmrho + \frac{\theta - \sin\theta}{\theta^3}\left(\hat\bmtheta\hat\bmrho + \hat\bmrho\hat\bmtheta + \hat\bmtheta\hat\bmrho\hat\bmtheta\right)\nonumber\\ 
    &-\frac{1-\frac{\theta^2}{2}-\cos\theta}{\theta^4}\left(\hat\bmtheta\hat\bmtheta\hat\bmrho+\hat\bmrho\hat\bmtheta\hat\bmtheta - 3\hat\bmtheta\hat\bmrho\hat\bmtheta \right)\nonumber\\
    &-\left(\frac{1-\frac{\theta^2}{2}-\cos\theta}{\theta^4} - 3\frac{\theta-\sin\theta-\frac{\theta^3}{6}}{\theta^5}\right)\hat\bmtheta\hat\bmrho\hat\bmtheta\hat\bmtheta
    \nonumber
\end{align}
Inversion of the matrix in \eqref{eq:barfoot_2014} gives the expression 
\begin{align}\label{eq:bullo95_3}
    \bmPhi^{-1}_r(\ad(\bmxi)) = \bmat \bmPsi^{-1}_{r}(\hat{\bmtheta}) & \bm0 \\ \bmC_r & \bmPsi^{-1}_{r}(\hat{\bmtheta}) \emat
\end{align}
for the inverse of the right Jacobian, as presented in \cite{Bullo1995}, where the submatrix $\bmC_r$ was unspecified. The expression 
\begin{align}\label{BarfootExressionC_r}
\bmC_r = -\bmPsi^{-1}_{r}(\hat{\bmtheta})\bmQ_r^\tr\bmPsi^{-1}_{r}(\hat{\bmtheta})
\end{align}
for this submatrix was obtained in \cite{Barfoot2014}. It is noted that a closed form solution for $\bmC_r$ was not found, which means that a closed form solution for the inverse of the right Jacobian in $SE(3)$ has not been reported so far. 

\subsection{Closed form for inverse Jacobians in SE(3)}

In this section we will derive a simple closed form solution for the inverse of the right and left Jacobians in $SE(3)$. In \cite{Bullo1995} it was shown that the inverse right Jacobian on $SE(3)$ can be computed as
\begin{align}\label{eq:bullo95_1}
    \bmPhi^{-1}_r 
    &= \bmI + \frac{1}{2}\ad(\bmxi) 
    + \gamma_1(\theta) \ad(\bmxi)^2 + \gamma_2(\theta) \ad(\bmxi)^4
\end{align}
where 
\begin{align}
    \gamma_1(y) &= (4-3\alpha(y)-\beta(y))/(2y^2) 
    \label{def_gamma1}\\
    \gamma_2(y) &= (2 - \alpha(y) - \beta(y))/(2y^4)
    \label{def_gamma2}\\
    \alpha(y) &= (y/2) \cot (y/2) \\
    \beta(y) &= (y/2)^2/\sin^2(y/2)
\end{align}
This was used in \cite{Bullo1995} to derive (\ref{eq:bullo95_3}). 

We will now derive a closed form solution for $\bmC_r$, which is a novel contribution. First it is observed that it follows from \eqref{AdjointMapMatrixSE3} that  
\begin{align}\label{eq:adxik=}
    \textbf{ad}(\bmxi)^k = \bmat \hat{\bmtheta}^k & \bm0\\ \bms_k & \hat{\bmtheta}^k \emat
\end{align}
where
\begin{equation}\label{sn_sum}
    \bms_k = \sum_{i=0}^{k-1} \hat{\bmtheta}^{k-i-1}\hat{\bmrho}\hat{\bmtheta}^{i}
\end{equation}
Next, it is seen from \eqref{eq:bullo95_3} and \eqref{eq:bullo95_1} in combination with \eqref{eq:adxik=} and \eqref{sn_sum}, that the matrix $\bmC_r$ must be of the form
\begin{equation}\label{eq:Cr=gammas124}
    \bmC_r = \frac{1}{2} \bms_1 + \gamma_1(\theta) \bms_2 + \gamma_2(\theta) \bms_4
\end{equation}
where $ \bms_1 = \hat\bmrho$, $\bms_2 = \hat\bmtheta\hat\bmrho + \hat\bmrho\hat\bmtheta$ and $\bms_4 = \hat\bmtheta^3\hat\bmrho + \hat\bmtheta^2\hat\bmrho\hat\bmtheta
    + \hat\bmtheta\hat\bmrho\hat\bmtheta^2 + \hat\bmrho\hat\bmtheta^3$. The expression for $\bms_4$ is simplified by observing that $\hat\bmtheta^2\hat\bmrho\hat\bmtheta + \hat\bmtheta\hat\bmrho\hat\bmtheta^2 = \hat\bmtheta(\hat\bmrho\hat\bmtheta + \hat\bmtheta\hat\bmrho)\hat\bmtheta = -2(\bmtheta^\tr\bmrho)\hat\bmtheta^2$. 
This in combination with $\hat{\bmtheta}^3 = -\theta^2\hat{\bmtheta}$ gives 
\begin{align}
   \bms_4 = 
   -\theta^2\bms_2 - 2(\bmtheta^\tr\bmrho)\hat\bmtheta^2 
\end{align}
Insertion of this expression for $\bms_4$ in \eqref{eq:Cr=gammas124} along with 
the expressions for $\gamma_1(\theta)$ and $\gamma_2(\theta)$ from (\ref{def_gamma1}) and (\ref{def_gamma2}) gives the closed form solution
\begin{align}\label{eq:final_Cr}
  \bmC_r  = \frac{1}{2}\hat{\bmrho} 
  + \frac{1 - \alpha(\theta)}{\theta^2}(\hat{\bmtheta}\hat{\bmrho}+ \hat{\bmrho}\hat{\bmtheta}) + \frac{\alpha(\theta) + \beta(\theta)-2}{\theta^4} (\bmtheta^\tr\bmrho)\hat\bmtheta^2
\end{align}
This gives the desired closed form solution for the inverse of the right Jacobian by inserting \eqref{eq:invrightjac_so3} and \eqref{eq:final_Cr} into  \eqref{eq:bullo95_3}. The closed form solution for the inverse of the left Jacobian is then
\begin{align}
    \bmPhi^{-1}_l(\ad(\bmxi)) = \bmPhi^{-1}_r(-\ad(\bmxi)) 
    = \bmat \bmPsi^{-1}_{l}(\hat{\bmtheta}) & \bm0 \\ \bmC_l & \bmPsi^{-1}_{l}(\hat{\bmtheta}) \emat
\end{align}
where $\bmC_l$ is equal to $\bmC_r$ except for a change of sign for the $\frac{1}{2}\hat{\bmrho}$ term. 

It is seen from the Taylor series expansions 
\begin{align}
\label{eq:Taylor(1-alpha(y))/y^2}
    \frac{1 - \alpha(y)}{y^2} &= \frac{1}{12} + \frac{y^2}{720} + \frac{y^4}{30\,240} + \ldots \\
    \frac{\alpha(\theta) + \beta(\theta)-2}{\theta^4} 
    &= -\frac{1}{720} - \frac{y^2}{15\,120} - \frac{y^4}{403\,200} + \ldots  \label{eq:lim_g2} 
\end{align}
that the coefficients in (\ref{eq:final_Cr}) are well-behaved for all $\theta$. 

%% file: LieAlgUKF.tex
\section{The Lie Algebraic UKF on SE(3)}\label{sec:UKF}


\subsection{System Dynamics}
The state is given by $\bmT \in SE(3)$, and the system dynamics is the kinematic differential equation 
\begin{equation}\label{System_Dynamics_continuous_SE(3)}
    \dot{\bmT} = \bmT[\bmV + \bmw]_{SE(3)}^\wedge
\end{equation}
where $\bmV$ is the vector form of the right velocity, and $\bmw \sim\mathcal{N}(\bm0,\bmQ)$ is a noise vector.

A discrete-time model is formulated, and the time propagation from time $t_k$ to $t_{k+1}$ is described by 
\begin{equation}\label{KinDiffEqSE3Log}
    \bmT_{k+1} = \bmT_{k} \exp([\bmxi_{k+1}]_{SE(3)}^\wedge)
\end{equation}
This means that the global state is given by the homogeneous transformation matrix $\bmT$, while the increment from one time instant to the next is described by the logarithm of the increment. This technique is similar to the usual formulation for multiplicative Kalman filters on the quaternions where the quaternion gives the global state, and a 3-dimensional vector is used in the update \cite{Crassidis2003}. The system dynamics of the logarithm, which is equivalent to the system dynamics (\ref{System_Dynamics_continuous_SE(3)}), is given by
\begin{equation}\label{System_Dynamics_continuous_log}
    \dot\bmxi = \bmPhi_{r}^{-1}\left(\textbf{ad}(\bmxi)\right) (\bmV + \bmw)
\end{equation}
This is discretized with the first order Euler method, which gives
\begin{align}\label{eq:disc_kin_on_log}
    \bmxi_{k+1} = \bmxi_k + h \bmPhi_{r}^{-1}\left(\textbf{ad}(\bmxi_k)\right) (\bmV_k + \bmw_k)
\end{align}
where $\bmw_k \sim\mathcal{N}(\bm0,\bmQ_k)$. 

The main difference to previous work is that the kinematic differential equation (\ref{System_Dynamics_continuous_log}) for the logarithm is used to describe the system dynamics, whereas previous work has used the kinematic differential equation (\ref{System_Dynamics_continuous_SE(3)}) on $G$. 

\subsection{System Measurements}

It is assumed that we can measure the full pose, and that the measurements are given by
\begin{align}\label{eq:fullposemeasurement}
    \bmZ = \bmT \exp\left([\bmnu]^\wedge_{SE(3)}\right) \ \in SE(3)
\end{align}
where $\bmnu_k \sim \mathcal{N}(\bm 0, \bm N_k)$ is the measurement noise vector. 

\subsection{Filter Dynamics}
\begin{figure}
    \centering
    \includegraphics[width=0.9\textwidth]{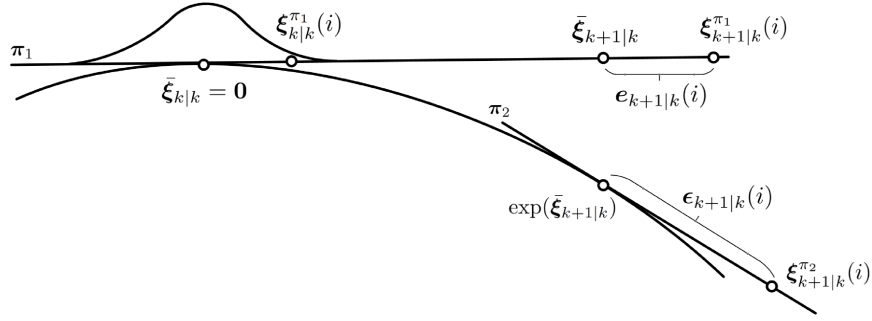}
    \caption{The relation between a predicted error vector $\bme_{k+1|k}(i)$ propagated on $T_{T_{k|k}}G$, denoted $\bm\pi_1$, and the corresponding error vector $\bmepsilon_{k+1|k}(i)$ on $T_{T_{k+1|k}}G$, denoted $\bm\pi_2$, is given through the Jacobian.}
    \label{fig:jac_vs_ad}
\end{figure}

The sigma points of the time update are given by  
\begin{align}
    \bmxi^a_{k|k}(i) = \bmat \bmxi^x_{k|k}(i) \\ \bmxi^w_{k|k}(i) \emat \in\mathbb{R}^{12}, \quad  i = 0,\ldots,2m
\end{align} 
which are computed according to step 4 and 5 in Algorithm \ref{alg:TU}. The vector $\bmxi^x(i)\in\mathbb{R}^6$ is related to the state variables, and $\bmxi^w(i)\in\mathbb{R}^{6}$ correspond to the process noise. The sigma points are propagated by using to the discretized dynamics in \eqref{eq:disc_kin_on_log}, which gives
\begin{align}\label{KinDiffEqSE3LogSigmaPoints}
    \bmxi^x_{k+1|k}(i) =& \bmxi^x_{k|k}(i) + h \bmPhi_{r}^{-1}\left(\textbf{ad}(\bmxi^x_{k|k}(i))\right) \left(\bmV_{m,k}+\bmxi^w_{k|k}(i)\right)
\end{align}
where $\bmV_{m,k}$ is the measured velocities. The predicted mean of the logarithm is computed as the weighted sum
\begin{align}\label{eq:sigmapointmeanvalue}
    \Bar{\bmxi}_{k+1|k} &= \sum^{2q}_{i=0} w^\mu_i \bmxi^x_{k+1|k}(i) \quad \in \mathbb{R}^{6}
\end{align}
where the weighting factor $w^\mu_i$ is computed according to step 2 in Algorithm \ref{alg:TU}. This is used to calculate the predicted mean of the global state as 
\begin{align}\label{eq:aprioristate}
    \bmT_{k+1|k} = \bmT_{k|k}\exp\left([\bar{\bmxi}_{k+1|k}]^\wedge_{SE(3)}\right) 
\end{align}
The propagated sigma points are written 
\begin{equation}
    \bmxi^x_{k+1|k}(i) = \bar{\bmxi}_{k+1|k} + \bme_{k+1|k}(i)
\end{equation}
Then, as in (\ref{exp_xi_with_nonzero_mean_BCH}), a first order approximation 
\begin{equation}
    \exp\left(\bar{\bmxi}_{k+1|k} + \bme_{k+1|k}(i)\right) =  \exp\left(\bar{\bmxi}_{k+1|k}\right)\exp\left(\bmepsilon_{k+1|k}(i)\right) \nonumber
\end{equation}
is then used where 
\begin{equation}
    \bmepsilon_{k+1|k}(i) = \Phi_r\left(\ad(\bar{\bmxi}_{k+1|k})\right)\bme_{k+1|k}(i)
\end{equation}
Figure \ref{fig:jac_vs_ad} illustrates how an error vector on the logarithm, $\bme$, maps to an error vector, $\bmepsilon$, on the tangent plane shifted by the exponential map of the mean vector, $\bar\bmxi$. The covariance is then calculated as
\begin{equation}
    \bmP_{k+1|k} = \sum_{i=0}^{2m} w_i^P \bmepsilon_{k+1|k}(i)\bmepsilon_{k+1|k}(i)^\tr 
\end{equation}
where $w^P_i$ is computed according to step 2 in Algorithm \ref{alg:TU}.

The new contribution is that the sigma points are computed as in (\ref{KinDiffEqSE3LogSigmaPoints}) from the kinematic differential equation of the logarithm. Then the mean of the time update can be computed as an average of vectors as in (\ref{eq:sigmapointmeanvalue}), while the covariance is calculated in terms of vectors on the tangent plane and is transformed to the next tangent plane by a matrix operation. This is computationally more efficient than averaging on $G$, which is typically done in previous work \cite{Hertzberg2013,Hauberg2013,Magalheas2018,Menegaz2018}, or computing the time propagation on $G$ and then averaging the logarithm and using parallel transport of the covariance as in \cite{Loianno2016}. In \cite{Hauberg2013} it was stated that the mean and the covariance can be computed in terms of the logarithm, and that the covariance could be transformed with parallel transport, however, details on the time propagation of the sigma points and the parallel were not included. In \cite{Brossard2017} the mean was not calculated from the sigma points, instead, based on \cite{Barfoot2014}, velocity measurements were used to calculate the mean. Moreover, we would like to point out that the proposed formulation gives a UKF for Lie groups that is more similar to the original formulation on $\mathbb{R}^n$, which could facilitate ease of understanding and implementation.

\begin{algorithm}
  \caption{UKF-LieAlg Time Update on $SE(3)$}\label{alg:UKF_prediction1}
  \begin{algorithmic}[1]
  \Statex \textbf{Input:} $\bmT_{k|k}, \bmP_{k|k}, \bmQ_k, \bmV_k, h, \alpha, \beta$
  \State $\lambda = (\alpha^2 - 1)m$ \Comment{$m = 12$}
  \State $w^\mu_0 = \frac{\lambda}{\lambda + m}, \ w^\mu_{j>0} = \frac{1}{2(\lambda + m)}$
  \Statex $w^P_0 = \frac{\lambda}{\lambda + m} + 1 -\alpha^2 + \beta, \ w^P_{j>0} = \frac{1}{2(\lambda + m)}$ \Comment{$\beta = 2$}
  \State $\bmP^a_{k|k} = \text{diag}(\bmP_{k|k}, \bmQ_k)$
  \State $\bmsigma_{k|k} = \Chol((m+\lambda)\bmP^a_{k|k})$
  \State $\bmxi_{k|k}^a(0) = \bm 0 \in \mathbb{R}^m $
  \Statex $\bmxi_{k|k}^a(i) = \text{col}_i(\bmsigma_{k|k}) $ \Comment{$i = 1 \ldots m$}
  \Statex $\bmxi_{k|k}^a(i + m) = -\text{col}_i(\bmsigma_{k|k}) $ 
  \State $\bmxi^a_{k|k}(i) = [ \bmxi^{x}_{k|k}(i)^\tr, \bmxi^{w}_{k|k}(i)^\tr ]^\tr$  \Comment{$\bmxi^x(i), \bmxi^w(i) \in \mathbb{R}^6$}
  \State $\bmxi^x_{k+1|k}(i) = \bmxi^x_{k|k}(i) + h\bmPhi_r^{-1}(\textbf{ad}(\bmxi^x_{k|k}(i)))(\bmV_k + \bmxi^{w}_{k|k}(i))$
  \State $\bar\bmxi_{k+1|k} = \sum_{i=0}^{2m}w_i^\mu \bmxi^x_{k+1|k}(i) $
  \State $\bme_{k+1|k}(i) = \bmxi^x_{k+1|k}(i) - \bar\bmxi_{k+1|k}$
  \State $\bmP_{k+1|k} = \bmJ_1 \left[\sum_{i=0}^{2m}w_i^P \bme_{k+1|k}(i)\bme_{k+1|k}(i)^\tr\right]\bmJ_1^\tr$  \Comment{$\bmJ_1 = \bmPhi_r(\textbf{ad}(\bar\bmxi_{k+1|k}))$}
  \State $\bmT_{k+1|k} = \bmT_{k|k} \exp([\bar\bmxi_{k+1|k}]^\wedge_{SE(3)})$
  \Statex \textbf{Output:} $\bmT_{k+1|k}, \bmP^{\pi_2}_{k+1|k}$
\end{algorithmic}
\label{alg:TU}
\end{algorithm}

\subsection{Measurement Update}

The measurement update is to a large extent based on the formulation of \cite{Brossard2017}. The sigma points of the measurement update are given by 
\begin{equation}
    \bmxi^a_{k+1|k}(i) = \bmat \bmxi^{x}_{k+1|k}(i) \\ \bmxi^{\nu}_{k+1|k}(i) \emat
\end{equation}
which are computed as in Algorithm~2. 
The measurements corresponding to the sigma points are
\begin{align}\label{eq:predicted_measurement}
    \bmZ_{k+1|k} = \bmT_{k+1|k}\exp\left([\bmzeta_{k+1|k}(i)]^\wedge_{SE(3)}\right) 
\end{align}
It is seen that the logarithms $[\bmzeta_{k+1|k}(i)]^\wedge_{SE(3)}$ are on the tangent plane at $\bmT_{k+1|k}$. The logarithm is given in terms of the sigma points as 
\begin{align}
    \exp\left([\bmzeta_{k+1|k}(i)]^\wedge_{SE(3)}\right) =\exp\left([\bmxi^x_{k+1|k}(i)]^\wedge_{SE(3)}\right)\exp\left([\bmxi^\nu_{k+1}(i)]^\wedge_{SE(3)}\right)
    \label{zeta_from_xi_and_nu}
\end{align}
which can be calculated as the logarithm of the expression in (\ref{zeta_from_xi_and_nu}), or it can be approximated on the tangent plane as 
\begin{align}
\bmzeta_{k+1|k}(i) = \bmxi^x_{k+1|k}(i) + \bmxi^\nu_{k+1}(i)
\end{align}
which was pointed out in \cite{Brossard2017}. 
The predicted measurement is computed as 
\begin{equation}
     \bar\bmzeta_{k+1|k} = \sum_{i=0}^{2m}w_i^\mu \bmzeta_{k+1|k}(i)
\end{equation}
where $w^\mu_i$ is found in step 2 in Algorithm \ref{alg:MU}, and the covariance is 
\begin{equation}
    \bmP_{zz} = \sum_{i=0}^{2m} w_i^P \Delta\bmzeta_{k+1|k}(i)\Delta\bmzeta_{k+1|k}(i) 
\end{equation}
where $\Delta\bmzeta_{k+1|k}(i) = \bmzeta_{k+1|k}(i) - \bar\bmzeta_{k+1|k}$. The cross covariance is 
\begin{equation}
    \bmP_{xz} = \sum_{i=0}^{2m} w_i^P \bmxi_{k+1|k}(i)\Delta\bmzeta_{k+1|k}(i)
\end{equation}
where the coefficient $w^P_i$ is found in step 2 in Algorithm \ref{alg:MU}.
The Kalman gain is computed as $\bmK = \bmP_{xz}\bmP_{zz}^{-1}$, and the error between the state and the measurement is, similarly to \cite{Baldwin2007}, defined as $\bmT_{k+1|k}^\inv\bmZ_{k+1}$ which follows from the relation in Equation \eqref{eq:predicted_measurement}. 
The innovation term is defined as the vector form of the error logarithm
\begin{align}
    \bmeta_{k+1} = \left[\log(\bmT_{k+1|k}^\inv\bmZ_{k+1})\right]^\vee_{SE(3)}
\end{align}
If a measurement $\bmZ_{k+1}$ is available, then the correction term is found according to 
\begin{align}
    \bmm_{k+1} = \bmK_k \bmeta_{k+1}
\end{align}
and the estimate updated from the measurement is found from
\begin{equation}
    \bmT_{k+1|k+1} = \bmT_{k+1|k} \exp\left([\bmm_{k+1}]^\wedge_{SE(3)}\right)
\end{equation}
The updated covariance is 
\begin{equation}
    \bmP^{-}_{k+1|k+1} = \bmP_{k+1|k} - \bmK \bmP_{zz} \bmK^\tr
\end{equation}
This covariance is calculated on the tangent plane at the predicted state $\bmT_{k+1|k}$ about the mean logarithm $\bmm_{k+1}$.
The covariance must be transformed to the tangent plane at $\bmT_{k+1|k+1}$, which is done as in (\ref{exp_xi_with_nonzero_mean_BCH}). This gives 
\begin{equation}
    \bmP_{k+1|k+1} = \bmJ_2\bmP^{-}_{k+1|k+1} \bmJ_2^\tr
\end{equation}
where $\bmJ_2 = \bmPhi_r(\textbf{ad}(\bmm_{k+1}))$. It is noted that this transformation of the covariance was not performed in \cite{Brossard2017}, but appeared in \cite{Bourmaud13} for an EKF on Lie groups.

As for the time update, the calculations are done in terms of vectors in the tangent plane, which simplifies implementation, and can potentially reduce computational costs. This method was used in \cite{Brossard2017}, while \cite{Hauberg2013,Hertzberg2013,Magalheas2018,Menegaz2018} used Lie group elements which were transformed to the tangent plane.

\begin{algorithm}
  \caption{UKF-LieAlg Measurement Update on $SE(3)$}\label{alg:UKF_prediction2}
  \begin{algorithmic}[1]
  \Statex \textbf{Input:} $\bmT_{k+1|k}, \bmP_{k+1|k}, \bmZ_{k+1}, \bmN_k \in \mathbb{R}^{6 \times 6}$
  \State $\lambda = (\alpha^2 - 1)r$ \Comment{$r = 12$}
  \State $w^\mu_0 = \frac{\lambda}{\lambda + r}, \ w^\mu_{j>0} = \frac{1}{2(\lambda + r)}$ 
  \Statex $w^P_0 = \frac{\lambda}{\lambda + r} + 1 -\alpha^2 + \beta, \ w^P_{j>0} = \frac{1}{2(\lambda + r)}$ \Comment{$\beta = 2$}
  \State $\bmP^a_{k+1|k} = \text{diag}(\bmP_{k+1|k}, \bmN_k)$
  \State $\bmsigma_{k+1|k} = \Chol((r+\lambda)\bmP^a_{k|k})$
  \State $\bmxi_{k+1|k}^a(0) = \bm 0 \in \mathbb{R}^r $
  \Statex $\bmxi_{k+1|k}^a(i) = \text{col}_i(\bmsigma_{k+1|k}) $ \Comment{$i = 1 \ldots r$}
  \Statex $\bmxi_{k+1|k}^a(i + r) = -\text{col}_i(\bmsigma_{k+1|k}) $   
  \State $\bmxi^a_{k+1|k}(i) = \bmat \bmxi^{x}_{k+1|k}(i)^\tr, \bmxi^{\nu}_{k+1|k}(i)^\tr \emat^\tr$\Comment{$\bmxi^{\nu}\in\mathbb{R}^{6}$}
  \State $\bmzeta_{k+1|k}(i) =\left[\log(\exp([\bmxi^x_{k+1|k}(i)]^\wedge_{SE(3)})\exp([\bmxi^\nu_{k+1}(i)]^\wedge_{SE(3)}))\right]_{SE(3)}^\vee$
  \State $\bar\bmzeta_{k+1|k} = \sum_{i=0}^{2m}w_i^\mu \bmzeta_{k+1|k}(i) $
  \State $\Delta\bmzeta_{k+1|k}(i) = \bmzeta_{k+1|k}(i) - \bar\bmzeta_{k+1|k}$
  \State $\bmP_{xz} = \sum_{i=0}^{2m}w_i^\mu(\bmxi_{k+1|k}(i))(\Delta\bmzeta_{k+1|k}(i))^\tr$
  \State $\bmP_{zz} = \sum_{i=0}^{2m}w_i^\mu(\Delta\bmzeta_{k+1|k}(i))(\Delta\bmzeta_{k+1|k}(i))^\tr$
  \State $\bmP^{-}_{k+1|k+1} = \bmP_{k+1|k} - \bmK \bmP_{zz} \bmK^\tr$ \Comment{$\bmK = \bmP_{xz}\bmP_{zz}^{-1}$}
  \State $\bmeta_{k+1} = \left[\log(\bmT_{k+1|k}^\inv\bmZ_{k+1})\right]^\vee_{SE(3)}$
  \State $\bmm_{k+1} = \bmK\bmeta_{k+1}$  
  \State $\bmP_{k+1|k+1}  = \bmJ_2\bmP^{-}_{k+1|k+1} \bmJ_2^\tr$ \Comment{$\bmJ_2 = \bmPhi_r(\textbf{ad}(\bmm_{k+1}))$}
  \State $\bmT_{k+1|k+1} = \bmT_{k+1|k} \exp([\bmm_{k+1}]^\wedge_{SE(3)})$ \label{StateUtdateT}
  \Statex \textbf{Output:} $\bmT_{k+1|k+1}, \bmP_{k+1|k+1}$
\end{algorithmic}
\label{alg:MU}
\end{algorithm}

%% file: ExperimentandSimulation.tex
\section{Simulations}
In this section, we present a comparison of 3 UKF filters: The UKF-LG of \cite{Brossard2017}, our proposed Lie algebraic UKF (UKF-LA) described in \ref{sec:UKF}, and the our method with optimization on the manifold in the prediction step (UKF-LA-Opt) as described in Section \ref{sec:calc_of_mean_manton}. The parameter $\alpha = 10^{-3}$ was used in all cases. The velocity measurements were body-fixed and obtained at a rate of 100 Hz, while the pose measurements were given in the inertial frame and obtained with a sample rate of 1 Hz. Furthermore, the measurements were elements of $SE(3)$, and the measurement noise was multiplicative and assumed to be given as in \eqref{eq:fullposemeasurement}. The angular and linear velocity in the time interval $t \in [0, T]$ were given by
\begin{align}\label{eq:velovectors}
    \bmomega^b(t) =& \bmat 0 & 0 & \frac{2t}{\frac{1}{1000}t^2+8t + 1} \emat^\tr \text{rad}/\text{s} \\ \label{eq:omegavectors}
    \bmv^b(t) =& \bmat \frac{t}{1+2t} & 0 & 0 \emat^\tr \text{m}/\text{s}
\end{align}
and were used to describe the right velocity as given in \eqref{eq:angvel_and_linvel_stacked}. The analytic expressions allowed for exact solutions, and we could therefore use an exact true trajectory for comparisons. All simulations were initiated with the covariance
\begin{align}
    \bmP_{0|0} = 0.01 \bmI_{6\times 6}
\end{align}
The process noise matrix was given as
\begin{align}
    \bmQ = \text{diag}(\bmQ_\omega, \bmQ_v)
\end{align}
where $\bmQ_\omega = \sigma_\omega^2 \bmI_{3\times 3}$ and $\bmQ_v = \sigma_v^2 \bmI_{3\times 3}$, which describes the uncertainty of the velocity measurements. The noise parameters were set to $\sigma_\omega = 0.1$ rad$/$s,  and $\sigma_v = 0.05$ m$/$s.

The covariance matrix describing the measurement noise matrix was given as
\begin{align}
    \bmN = \text{diag}(\bmN_R, \bmN_p)
\end{align}
where $\bmN_R = \sigma_R^2 \bmI_{3\times 3}$ and $\bmN_p = \sigma_p^2 \bmI_{3\times 3}$. The measurement noise parameters were given as $\sigma_R = \frac{\pi}{180}$~rad and $\sigma_p = 0.01$~m. The angular error at time $t_k$ was computed as 
    $\theta_{e,k} =\| [ \log(\bmR_{k|k}^\tr\bmR_{k})]^\vee_{SO(3)}\|$
where $\bmR_{k|k}$ is the estimated rotation matrix between the body-fixed frame and the inertial frame, and $\bmR_{k}$ was the true attitude of the system. 
The positional error was found as
    $r_{e,k} = \| \bmr_{k|k} - \bmr_{k} \|$
where $\bmr_{k|k}$ was the estimated position and $\bmr_{k}$ was the true position, both given in the inertial frame.


\begin{figure}[H]
    \centering
    \includegraphics[width=0.6\textwidth]{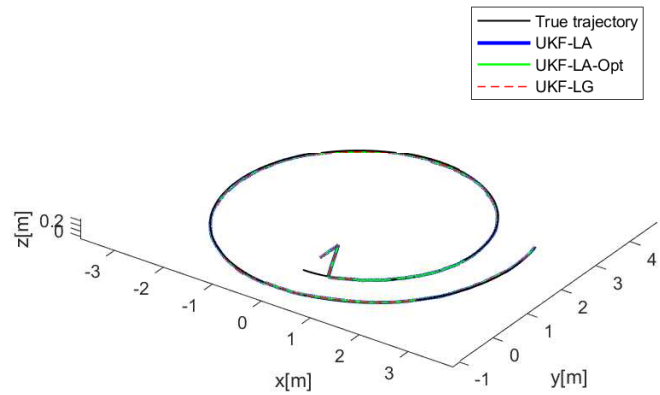}
    \caption{The body frame was initiated with a $\ang{45}$ angular error as well as $0.5$ meters off the initial position. Before any measurement updates had been obtained the predicted motion was incorrect and the body frame was heading away from the $xy$ plane which the true trajectory reside on, but was able to converge towards the true trajectory as the pose was measured.}
    \label{fig:spiral_with_offset}
\end{figure}

\begin{figure}[H]
    \centering
    \includegraphics[width=0.6\textwidth]{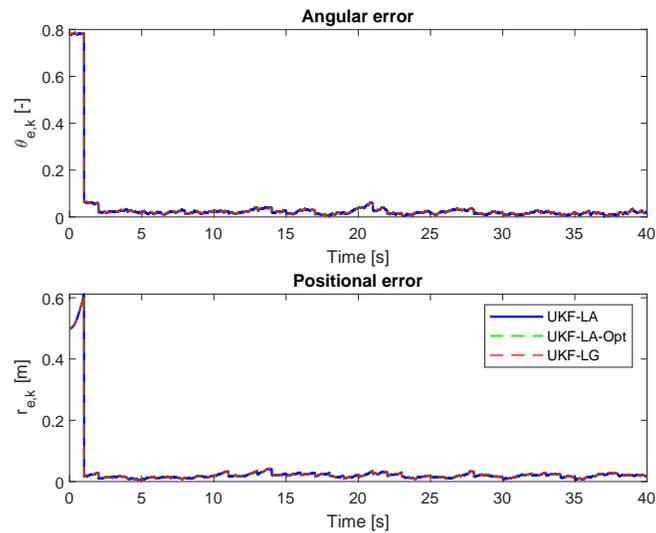}
    \caption{The estimation error due to a poorly chosen initial condition was reduced when the pose of the body had been measured.}
    \label{fig:spiral_with_offset_error}
\end{figure}

\subsection{Case Studies}
The first case the trajectory to be estimated was in terms of the the right velocity in \eqref{eq:velovectors} and \eqref{eq:omegavectors} over $T = 40$ seconds. The trajectory is in the $xy$ plane. The estimator had initial state given by
\begin{align}
\bmT_{0|0} = 
\bmat 
\sqrt{2}/2 & 0 & -\sqrt{2}/2  & 0  \\
0 & 1 & 0 & 1/2  \\
\sqrt{2}/2 & 0 & \sqrt{2}/2  & 0  \\
0 & 0 & 0 & 1  \\
\emat 
\end{align}
which corresponds to an initial estimation offset of a rotation of $\ang{45}$ about the $y$ axis, and a positional offset of $0.5$~m along the $y$ axis. Due to the initial angular offset, estimated state left the $xy$ plane for the first 100 samples (Figure~\ref{fig:spiral_with_offset}), until the first pose measurement made the estimator errors converge to values close to zero, as seen in Figure~\ref{fig:spiral_with_offset_error}.

\begin{figure}[H]
    \centering
    \includegraphics[width=0.6\textwidth]{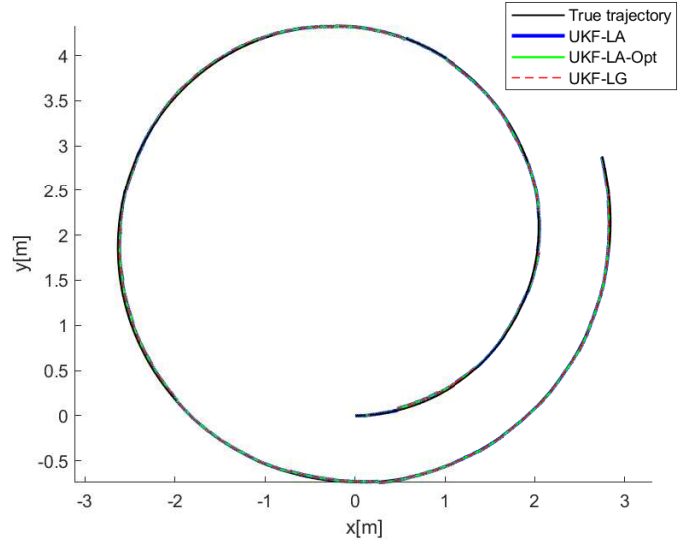}
    \caption{Spiral trajectory initiated with zero offsets.}
    \label{fig:spiral_no_offset}
\end{figure}

\begin{figure}[H]
    \centering
    \includegraphics[width=0.6\textwidth]{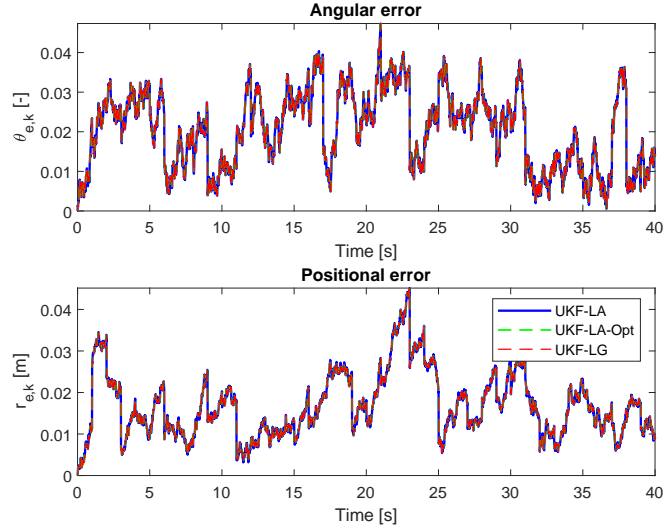}
    \caption{Angular and positional error compared with the ground truth when the system after the system had been initialized with zero offsets.}
    \label{fig:spiral_no_offset_error}
\end{figure}

In the second case, the three estimators were had initial states given by identity matrices, such that $\bmT_{0|0} = \bmI_{4\times 4}$ before estimating the trajectory. It is seen in Figure \ref{fig:spiral_no_offset} that the all the three estimators tracked the trajectory with high accuracy. 
The estimates provided by the three filters were close to indistinguishable when they are used with the same set of measurements. This is seen in Figures~\ref{fig:spiral_with_offset_error} and \ref{fig:spiral_no_offset_error}.

\subsection{Computational efficiency}

In order to evaluate the computational effort of the UKF filters, 100 different sets of full pose measurements of 1000 samples were generated, where the time update was performed at a rate of 100 Hz, and the measurement update was performed at 1 Hz. The computational time over each set of measurements was evaluated for each estimator, and the mean computational time was computed. The average computational time spent for each estimator is provided in Table~\ref{tab:comptime} together with the difference given in terms of percentages.
\begin{table}
\begin{center}
 \begin{tabular}{||c c c||} 
 \hline
 UKF-LA & UKF-LG & UKF-LA-Opt \\ [0.5ex] 
 \hline\hline 
 0.6117 s & 1.6385 s & 1.9220 s   \\
 37.3 \% & 100 \% & 117.3 \%  \\ 
 \hline
\end{tabular}
\end{center}
\caption{The UKF-LA required less computational effort compared to UKF-LG and UKF-LA-Opt.}\label{tab:comptime}
\end{table}
It is seen that the proposed UKF-LA gave  computational time which was 37.3 \% of the computational time of UKF-LG, which was set to 100 \%. When the mean was obtained through optimization on the manifold, then UKF-LA was slower than UKF-LG. It is noted that closed form solutions of the exponential and logarithmic maps on $SE(3)$, as described in Equation \eqref{eq:expmSE3_CF} and \eqref{eq:logmSE3_CF}, were used for efficiency in computation. If library functions in MATLAB were used for computation of the exponentials and the logarithms maps, then UKF-LA could perform up to 10 times faster than UKF-LG. This becomes evident when studying the prediction step presented in \cite{Brossard2017}, as the exponential map must be computed twice, the logarithmic map once, and two matrix multiplications are required for each sigma point in each prediction step. In contrast, the exponential map is computed once per prediction step in UKF-LA, and no logarithms must be called unless the estimated mean is obtained through optimization.

%% file: Conclusion.tex
\section{Conclusion and future works}
A UKF for matrix Lie groups has been proposed where the time propagation is formulated in terms of the kinematic differential equation of the logarithm. The proposed method is to a large extent formulated in terms of vector operations on the Lie algebra, and the formulation is closer to the original UKF on $\mathbb{R}^n$ than previous works on Lie groups. This leads to efficient formulations and potentially to reduced computational costs, in particular in the time update. The paper includes details on how to implement the proposed UKF for $SE(3)$. The method was compared to the UKF-LG of \cite{Brossard2017} in simulations on $SE(3)$, where it was found that the difference in the estimator errors was not significant, while the computational cost of the proposed method was 37 \% of the UKF-LG. The proposed method with averaging of the sigma points on the Lie algebra was compared to a method with averaging of the sigma points on the manifold. Again, there was no significant difference in the estimation error, and the computation time of the proposed method was 32 \% of the method with averaging on the manifold. Future work may include equations of motion and different sensor systems including IMUs and bias modeling.